# Some considerations on how the human brain must be arranged in order to make its replication in a thinking machine possible

Emanuel Diamant
VIDIA-mant, POB 933 Kiriat Ono, 55100 Israel
<emanl.245@gmail.com; http://www.vidia-mant.info>

**Abstract:** For the most of my life, I have earned my living as a computer vision professional busy with image processing tasks and problems. In the computer vision community there is a widespread belief that artificial vision systems faithfully replicate human vision abilities or at least very closely mimic them. It was a great surprise to me when one day I have realized that computer and human vision have next to nothing in common. The former is occupied with extensive data processing, carrying out massive pixel-based calculations, while the latter is busy with meaningful information processing, concerned with smart objects-based manipulations. And the gap between the two is insurmountable.
To resolve this confusion, I had had to return and revaluate first the vision phenomenon itself, define more carefully what "visual information" is and how to treat it properly. In this work I have not been, as it is usually accepted, "biologically inspired". On the contrary, I have drawn my inspirations from a pure mathematical theory, the Kolmogorov's complexity theory. The results of my work have been already published elsewhere. So the objective of this paper is to try and apply the insights gained in course of this my enterprise to a more general case of information processing in human brain and the challenging issue of human intelligence.
In the case of vision, the key point of my approach was: information is a linguistic description, a letter-based sequence, a string, a piece of text. Then, two types of information can be defined: physical information and semantic information. Both have to be represented as a top-down coarse-to-fine evolving hierarchy with the most simplified description at the top, and most detailed descriptions at the bottom. The lowest level of a semantic hierarchy is actually a set of physical attributes related to the lowest semantic items. Thus, an interpretation of physical information (which is delivered by sense organs) comes as a result of establishing a fitting similarity (an association) between the current physical information (at the system's input) and the physical information retained at the attribute level of a particular semantic description.
Applying these principles to information processing in human brain, we can immediately notice: bursts of spiking neuronal pulses cannot be accepted as the right means of information exchange between neurons. Molecularly-encoded information texts (strings) have to be recognized as the right means of information transfer between brain units. (Indeed, there are now plenty of evidence that astrocytes communicate with neurons without any pulse trains).
Next: To process semantic information a system has to possess a memory, where descriptions of previous semantic hierarchies (previous events and experiences) are to be stored. From computer building know-how we know that there must be two sorts of memories in a system – a flexible operational (Read/Write Random Access) memory and a rigid (Read Only, hardware) memory. If this is not an artifact, it must hold also for a natural biological information processing systems. Indeed, two kinds of memory can be easily discerned in every biological system. The rigid one is the DNA code of a genome. The flexible one is the epigenetic memory. We can point at the plasmids as the location of an operational memory in bacteria. We can point at the dendritic spines as the location of an operational memory in mammals (including humans).
From visual information processing studies we have already learned that the memory content must always be brought into a memory from the outside. In the case of a genetic memory, its content is inherited from the parents. That is the so-called Vertical Information Transfer mode. In the case of an operational memory, its content is delivered in a Horizontal Information Transfer mode, by means of plasmids in bacteria, or verbal/written information exchange in humans. Since in bacteria the operational information in plasmids can be easily incorporated into (or extracted from) the main genome, it is reasonably to assume that they are similar in their structure. That is, the descriptions in the plasmids are written in the same manner (in the same proto-language) as the descriptions in the genome. The same arguments can be applied to the dendritic spines "stuffing". For the same reason, we can say that our thoughts are DNA-like memory strings fetched out from the operational memory (from the dendritic spines locations), and replicated in the same manner as a single gen is replicated. That is, one copy of a DNA double helix is left in the spine for the future use, and the other copy is sent to further processing stations. During the copy-making process, the same editing procedures are applied to the dispatched version of the string. These are similar to gene splicing and editing procedures in a gene replication process. In such a way, our thinking and operational memory refueling could be understood and materialized.
This understanding must shed a new light on the relations between natural and cultural evolution, (Vertical and Horizontal Information Transfer), on the extinction of Dawkins' memes, and on the crucial role that cultural evolution plays, reshaping our surrounding and accelerating the pace of our continuously growing intelligence.

# Некоторые размышления о том, как должен быть устроен человеческий мозг, чтобы по его образу и подобию можно было бы создавать думающие машины

Эммануэль (Амик) Диамант
VIDIA-mant, Israel
<emanl.245@gmail.com; http://www.vidia-mant.info>

## 1. Пролог

2006-й год был ко мне особенно неласков – из 13 статей, посланных на международные конференции этого года, 12 вернулись назад отвергнутыми. И не просто так – «отвергнуты»! Все, без исключения, – «Решительно отвергнуты!». Программная комиссия ECCV (Европейская Конференция по Компьютерному Зрению) была со мной особенно строга: «Это философская статья... Однако, ни в правилах ECCV, ни в её традициях нет места для такого рода статей. Весьма сожалеем, (Sorry)».

Боже милосердный! Как же такое может быть в твоём мире? Хотя, с другой стороны, неприязнь этих людей (моих анонимных рецензентов и членов всяческих программных комиссий) можно легко и понять – я ведь, действительно, во всех своих делах и писаниях стараюсь занять как можно более отстранённую, философскую позицию. Философия для меня не бранное слово. Философия для меня – это естественная потребность взглянуть на вещи с возможно более общей точки зрения, чтобы не потерять перспективы, более широкого взгляда на предмет обсуждения, и не пасть жертвой узких частных определений, ловушек локального поиска в ограниченном пространстве или алхимии бесконечного перебора случайных решений, подходящих только для очень ограниченного круга задач.

Особенно этот философский взгляд важен в таких, казалось бы, повседневных, общедоступных и обыденных вещах, как зрение. Эка невидаль – человеческое зрение! Любому доступно. Любой им владеет. Для любого, кажется, нет ничего проще: посмотрел, увидел и – всё понял... Однако же... Любые попытки перепоручить это дело машине (электронной видеокамере с приставленным к ней компьютером) вот уже полвека никому не удаются и повсеместно кончаются полным провалом. (Даже большие умники из ECCV нисколько в этом до сих пор не преуспели, хотя традиционно надуваются и пыжатся безмерно. Ещё бы – Европейское Сообщество тратит сегодня на эти вещи огромные деньги, больше, чем когда-то тратили США, готовясь высадить человека на Луну. Помните «Аполлон»?! Так сейчас это во много раз грандиознее, особенно по деньгам).

Традиционно считается, что качество картинки прямо связано с разрешающей способностью видео-системы – чем она выше, тем лучше мы видим. Чем больше деталей, тем выше качество изображения, тем оно (должно быть) ценнее для потребителя. Успехи современной электроники направлены именно в эту сторону: постоянное, максимально возможное увеличение разрешающей способности фотоприёмников – от дешёвых цифровых фотоаппаратов с матрицей в 5 - 8 Мегапикселей (пиксель – от английского «pixel», что значит «picture element», отдельный элемент картинки) через профессиональные камеры с разрешением до 16 Мегапикселей и до специальных видеокамер медицинского, военного и аэро-космического назначения с разрешением от 30 до 85 Мегапикселей. Разумеется, для отображения таких картинок требуются соответствующие экраны специального (высокого) разрешения. И, конечно же, такие экраны уже имеются как для специальных приложений (военных, медицинских), так и для ширпотреба: например, стандарт HDTV (Телевидение Высокого Разрешения), обязательный для всех телевизоров нового поколения, предусматривает разрешение в 1920x1080 пикселей (против обычного 720x576) или экраны для персональных компьютеров – 1280x1080 пикселей.

При этом, конечно, народная мудрость о том, что «видят не глазами, а умом», (в русском языке этимология слова «видеть» ведётся от слова «ведать», т.е., «знать», «понимать», а вовсе не от «вида» (что есть «форма», «размер» или что-то подобное), – всё это, конечно, в расчёт не принимается. А жаль. Потому что в отличие от машины человек видит не набор пикселей, а их связанные, осмысленные объединения, которые принято называть «образами», «объектами», «предметами».

Как и где эти наборы отдельных пикселей превращаются в целостные предметы, которые мы осмысленно воспринимаем и которыми мы манипулируем в нашем сознании? На этот вопрос ни у кого сегодня нет ответа. Т.е., на вопрос «где?» ответ есть: «в голове», «в уме». Но это мало чем кому-то может помочь. Потому что, что такое «ум» – тоже пока никто не знает. А тогда вопрос «как?» и приложить не к чему.

## 2. Ти ж мене підманула...

Пренебрежительное отношение к философским проблемам зрения мгновенно оборачивается для нас весьма чувствительными проблемами повседневного использования машинного или, как его ещё часто называют, компьютерного зрения, которое у нас в большом ходу и которое создано для того, чтобы максимально удовлетворить наши зрительные запросы и потребности.

Человек – существо, постоянно испытывающее информационный голод. «Infovore» (что значит: информационно-голодный) – так определил это Ирвинг Бидерман, один из отцов-основателей науки о компьютерном зрении (Biederman, 2006). Может быть, поэтому мы всегда так жадно всматриваемся в телевизор. Может, именно поэтому таким успехом пользуются у нас видеофоны и камерафоны (cameraphone – видеокамера, совмещённая с сотовым телефоном). Одна только Nokia в 2007 году продала 440 миллионов таких телефонов, т.е. мобильных телефонов оснащённых видеокамерами (Nokia, 2008). Это 40% мировых продаж, т.е. более миллиарда видеофонов было продано в одном только 2007 году. В 2009 году их будет продано ещё больше – больше, чем было изготовлено любого типа фотокамер за все годы с момента изобретения фотографии (Thevenin et al., 2008).

Всё вместе это привело к невиданному доселе потоку видеоинформации, бурлящему вокруг нас. Чтобы как-то справиться с этим потоком, срочно принимаются меры по стандартизации правил организации и регулирования этого потока, т.е. правил кодирования и обмена визуальной информации, циркулирующей в этом потоке.

Естественно, особенности человеческого зрения должны быть тут приняты во внимание. Сомнений по этому поводу ни у кого никогда не возникало – комиссия по разработке нового стандарта кодирования видеоизображений MPEG-4, разработка которого началась в 1994 году, а окончательная редакция была завершена в 1999 году, торжественно провозгласила: Новый стандарт будет объектно-ориентированным (т.е. кодироваться в нём будут не пиксели, а визуальные объекты). Для этой цели в стандарт был введён целый ряд новых понятий: VO (Visual Object ), VOP (Visual Object Plane), VOL (Visual Object Level). (Источник: Puri & Eleftheriadis, 1998).

Ввести-то их ввели, да только, как искать их и находить в картинке, – не сказали. Потому, что не знали. И до сих пор не знают. А потому всё дальнейшее развитие и усовершенствование стандарта (а за прошедшие 10 лет сделано тут совсем немало) пошло по пути негласного усовершенствования пиксель-ориентированного кодирования. Объектно-ориентированный MPEG-4 на самом деле есть модифицированный пиксель-ориентированный MPEG-2, хотя его стыдливо и переименовали в MPEG-4/10 Advanced Video Coding Standard (H.264/AVC). По сути это есть модифицированный пиксель-ориентированный MPEG-2. А у этого, разумеется, есть свои неотвратимые (однако, стандартом узаконенные и к повсеместному применению обязательные) печальные последствия.

По-пиксельная обработка картинки требует затрат времени и энергии. Особенно, если размеры картинки неустанно растут. Это привело к созданию совершенно новых типов обрабатывающих устройств, т.н. цифровых процессоров, предназначенных для максимально быстрой переработки данных в картинках. Таким, например, является Analog Devices TigerSHARC ADSP-TS201S процессор (TigerSHARC – по-английски Тигровая Акула), с производительностью в 3.6 GFLOPs (Гигафлоп – Гига (единица с девятью нулями) вычислительных (с плавающей запятой) операций в секунду). Для больших картинок (HDTV) и этого, конечно, мало. В этом случае компания BittWare предлагает PCI Mezzanine Card (специальную надстроечную плату), на которой располагаются сразу четыре TigerSHARCа. И это ещё не всё – в одном компьютере могут быть установлены до четырёх таких плат с общей производительностью в 57 GFLOPs (Bittware, 2007). Сколько ж при этом энергии тратится? Лучше не спрашивайте. Хотя на самом деле, кого это интересует?!.. Ведь такие решения предназначены для стационарных устройств, где проблемы энергопотребления (и охлаждения, в свой черёд) – мало кого волнуют.

Ну, а как же быть с передвижными устройствами? – спросите вы. – Как быть с нашими горячо любимыми мобильными видеофонами? Туда ведь даже один единственный TigerSHARC не всунешь?..

Ответ на этот вопрос, оказывается, удивительно прост. В то время как разрешение фотоприёмника вашего видеофона непрерывно растёт (о чём вам всё время навязчиво напоминает реклама) – от 1.5 Мегапикселей (1280x1024) до 5 (2580x1930), 8 (3264x2444), 12 (4220x2820) и наконец до 14 Мегапикселей (4570x3050) – сегодня уже есть и такие, – эффективный размер картинки, которую вы можете наблюдать и передавать дальше, меняется весьма незначительно: 80x60 пикселей, потом 160x120, и наконец 352x288 пикселей (в Америке – 352x240 пикселей) – это уже новейший Common Intermediate Format Standard (CIF). Вот так, уважаемые потребители. Не хотите, можете не покупать, вас ведь никто не заставляет – рынок свободный. (Да ещё какой рынок! Сотни миллиардов долларов в год! Программа «Аполлон» стоила всего 25 миллиардов, правда, в ценах почти 40-летней давности.)

Печальная повесть о том, как на этом свободном рынке законно надувают вольных и суверенных в своём выборе потребителей, на этом не кончается. Следующая история касается взрывного распространения систем видеонаблюдения для целей военной (оборонной), гражданской (общественной) и, конечно же, личной безопасности. Предполагается, что рост объёма продаж, связанных с установкой систем видеонаблюдения, вырастет с 4.9 миллиардов долларов в 2006 году до более, чем 9 миллиардов долларов в 2011 году (Video surveillance, 2007).

Считается, что движущей силой этого взрыва является широкое распространение и внедрение сетевого Интернет Протокола ИП (IP – Internet Protocol) – своеобразной технологии, позволяющей связать в одну единую систему рассеянные в пространстве (или по территории) видеокамеры и центр (центры) их компьютерного обслуживания. Одна только маленькая неувязка – эффективный размер картинки, которая предусмотрена Интернет Протоколом, определён CIF Стандартом в 352х288 пикселей. Мало, очень и очень мало. Особенно, если учесть, что картинки эти предназначены для рассматривания человеком. Ибо сегодня только человек, рассматривающий картинку, может понять, что же там (на картинке) есть или что же там происходит. А это значит, что для эффективной работы системы видеонаблюдения, к выходу каждой системы должен быть всегда (и обязательно) приставлен человек – 24 часа в сутки, 7 дней в неделю, 52 недели в году... Без этого ничего работать не будет, без этого никакой видеобезопасности вы не получите. А человек – очень хрупкое создание, больше пяти минут в пустой экран (т.е. экран, на котором ничего не происходит) смотреть не может, засыпает с открытыми глазами (так уж он физиологически устроен). Однако это мало кого волнует – бизнес есть бизнес, и спрос удовлетворяется предложением. Более того, чтобы сэкономить на рабочей силе, дежурному наблюдателю на один экран, как правило, выводятся 4, 8, и даже 16 камер (CIF очень даже в этих случаях уместен). Да и какая разница? Эффективность ныне действующих систем видеонаблюдения в любом случае весьма сомнительна.
(В 2002 году, я был приглашён как-то на обсуждение проекта системы видеозащиты одного очень важного (стратегически важного) объекта. Общая концепция проекта была уже кем-то заранее продумана, а поэтому основной проблемой, выставленной на обсуждение, было: сколько камер выводить на один дисплей? 16? 20? 24? Или 32? Когда обсуждение дошло до меня, я предложил весь этот проект закрыть (толку-то всё равно никакого), а сэкономленные деньги раздать университетам, чтобы они могли разрабатывать системы автоматического слежения и анализа содержания картинок. (Пригодится, если не сегодня, так завтра). В итоге – меня перестали приглашать на подобные обсуждения. А ту систему – воплотили и запустили! Она функционирует по сей день. С большой пользой (выгодой) для некоторых инициаторов проекта, и, разумеется, нулевой для страны и для общества.)

## 3. Ой, казала мені мати тай наказувала...

Создание видеосистем, способных без помощи или участия человека, самостоятельно понимать, что они видят, и осмысленно реагировать на увиденное (хотя бы на уровне привлечения внимания оператора к ситуации, представляющей возможную угрозу) на самом деле задача совсем не новая и точно уж не мной задуманная и поставленная. Начало этому было положено ещё на Дортмундской встрече в 1956 году (McCarthy, et al. 1955), где четверо отцов-основателей (МакКарти, Минский, Рочестер и Шеннон) предложили программу исследований в области, которая будет с тех пор называться Искусственным Интеллектом (Artificial Intelligence) и которая в течение полувека будет заниматься проблемами Думающих Машин (Thinking Machines), как их ещё до этого задумал и определил Алан Тюринг, основоположник вычислительной техники и будущих компьютеров (Turing, 1950).
Как связаны, спросите вы, системы видеонаблюдения и автоматического анализа увиденного с Думающими Машинами и Искусственным Интеллектом? В том-то и дело, что связаны, и даже очень. О том, что «видеть» это «ведать», вы уже слышали. А вот о том, что половина человеческого мозга занята переработкой визуальной информации, вам, конечно, интересно будет узнать (Milner & Goodale, 1998). Академик Репин считает даже, «что мозг человека на 80 процентов загружен зрительной информацией» (Репин, 2000). Во всяком случае, для отцов-основателей это выглядело абсолютно естественным – изучение проблем мозга нужно начинать с изучения проблем зрения. И никаких проблем в этом никто из них тогда не видел. Стив Гранд, например, выступая на Panel Discussion at the Artificial General Intelligence Workshop в 2006 году, рассказал такую историю: «У Родни Брукса (одного из корифеев Искусственного Интеллекта) хранится меморандум, представленный Мервином Минским (ещё один из основоположников, инициатор Дортмундской встречи), в котором он (Минский) предлагает (речь идёт о конце пятидесятых) запустить проект, который разрешит все уже возникшие к тому времени проблемы зрения, подрядив для этого на время летних каникул кого-нибудь из старших студентов. «Я не знаю, – говорит Стив Гранд, – где теперь этот студент. Но я думаю, что свою работу он до сих пор так и не закончил.» (Panel Discussion, 2006).
Зрение, вопреки всеобщему убеждению, вдруг оказалось чрезвычайно сложной проблемой. Да и многие другие исходные (начальные) предпосылки теории Искусственного Интеллекта вдруг оказались весьма сомнительными, отчего великолепное здание науки о Думающих Машинах начало буквально на глазах разрушаться и сыпаться, распадаясь на мелкие, частные суб-дисциплины. Так случилось, например, с Машинным Обучением. Первоначально считавшееся подразделом Искусственного Интеллекта, оно очень быстро выродилось в самостоятельную дисциплину. При этом точного определения, что такое Машинное Обучение и как оно соотносится с Искусственным Интеллектом, так до сих пор и не существует. Мартин Хьюттер в своей статье «Универсальный Интеллект: Определение Машинного Интеллекта» (Legg & Hutter, 2007) приводит 70 с лишним определений Машинного Интеллекта, взятых им из трудов ныне здравствующих и активно работающих авторитетов в этой области. Никакого согласия между этими определениями нет (и быть не может). Существующий разнобой можно объяснить только одним – отсутствием обобщающего философского подхода, наивной верой в то, что из разрозненных мелких кирпичиков может быть собрано великое здание Машинного Разума.

В условиях такого разнобоя центр тяжести научных исследований естественно смещался, и сместившись, постепенно уступил своё место изучению проблем так называемого «биологического зрения». Вдруг стало модным учиться у великого творца всего сущего в этом мире – у Природы. К концу 1970-х – началу 1980-х годов прошлого столетия оформилась и утвердилась в науке современная школа биологического зрения, которая на десятки лет вперёд определила пути прогресса во всех областях, связанных с изучением работы мозга. В том числе и с пониманием функций «компьютерного зрения» или Интеллекта Думающих Машин.

Ведущими тут оказались работы Давида Марра (Marr, 1978; Marr, 1982), Анны Трейзман (Treisman & Gelade, 1980) и Ирвинга Бидермана (Biederman, 1987). Нет смысла углубляться в детали их трудов, но буквально в двух словах следовало бы изложить их основную идею. Сводится она к следующему: Зрение (по Анне Трейзман) есть взаимодействие двух потоков обработки информации. Один из них «восходящий», «снизу-вверх» идущий (bottom-up directed) процесс (от глаз, через «нижние», ближайшие к глазам области мозга к «верхним», сознательным, думающим областям), в котором бессознательно выделяются и обрабатываются мелкие (пиксельные) элементы информации, которые объединяются в более крупные промежуточные элементы картинки (image features). Второй поток – «нисходящий», «сверху-вниз» направленный процесс (top-down directed), в котором эти первоначально выделенные элементы теперь уже осмысленно объединяются (группируются) в целостные объекты (предметы). Именно эти предметы мы и «видим», именно ими мы и оперируем, когда «рассматриваем» картинку, пытаясь понять, что же там изображено (Treisman & Gelade, 1980).

Принципы работы восходящего потока были с самого начала легко и интуитивно понятны каждому, довольно просто формализовались математически, а поэтому быстро и охотно были усвоены и подхвачены всеми, в том числе и разработчиками «машинного» или «компьютерного» зрения. С тех пор они только этим и занимаются, пытаясь все проблемы машинного зрения свести к решению этих «нижнеуровневых» задач (low-level image processing tasks, low-level bottom-up pixel-oriented image processing).

Принципы же работы нисходящего потока с самого начала были весьма туманны и расплывчаты. Понятно, что для успешного объединения разрозненных деталей в какое-то общее целое нужно какое-то особое изначальное знание принципов, по которым это объединение должно было бы проводиться.Понятно, что располагается это знание в самой «верхней», «сознательной» зоне мозга (и поэтому весь процесс считается направленным «сверху-вниз», поэтому-то он называется «нисходящим»).

Острейшая необходимость понять принципы, по которым должно осуществляться это объединение, привело к возникновению и интенсивной разработке целой новой области знания, получившей название Проблемы Связывания (The Binding Problem). Её решению было посвящено (и сегодня посвящается) масса средств и усилий. Многолетним напрасным трудам на этом поприще посвящён специальный выпуск журнала Neuron (Volume 24) за 1999 год, а также попытки самой Анны Трейзман как-то продвинуться в этом направлении (Treisman, 1996). Когда оказалось, что все эти усилия ни к чему не ведут, Кох и Крик (да, да, тот самый Крик, Нобелевский лауреат, который вместе с Уотсоном открыл двойную спираль ДНК) предложили идею «гомункулуса», маленького человечка, который сидит у нас в голове, и всё знает, всё за нас решает. В том числе и как разумно связывать разрозненные элементы картинки в целостные объекты и предметы (Crick & Koch, 2000).

Теория гомункулуса не прижилась. Но потребность в «связывании» оставалась и даже становилась всё более острой – без решения проблемы связывания невозможно было осуществить объектно-ориентированное описание картинок. А без этого не может быть решена главная проблема сегодняшнего Интернета – проблема «человекообразного» поиска в сети нужного материала по его визуальному содержанию. В специальной литературе это называется Поиск Изображений по Содержанию (CBIR - Content-Based Image Retrieval). Поскольку даже теории биологического зрения не могут объяснить, как из отдельных элементов картинки формируются осмысленные визуальные объекты, а без этого ни о каком осмысленном (семантическом) описании содержания картинки не может быть и речи, и следовательно, поиск по содержанию (по смыслу) становится бессмысленным. Тогда – зачем вообще нам все эти теории?!! Тогда – Даёшь алхимию! Даёшь единственно доступное и очевидно напрашивающееся решение – из обильной и доступной информации нижнего уровня будем добывать недоступную нам информацию верхнего уровня! (Чтобы вы не подумали, что я опять в очередной раз возвожу поклёп на добрых людей, сошлюсь лишь на несколько работ в этой области, выхваченных мною наугад: Mojsilovic & Rogowitz, 2001; Zhang & Chen, 2003; Itti, 2005; Serre et al., 2005; Hare et al., 2006; Kveraga et al., 2007. В действительности же список этот бесконечен.)

Глядя на весь этот праздник жизни, я только и могу сказать: «Мне очень жаль, но человеческое зрение и человеческое мышление устроены совсем не так и совсем не так работают...»

## 4. Путь далёк у нас с тобою...

Я не хочу сказать, что я всегда был такой «умный», как это может кому-нибудь показаться сегодня. Отнюдь. Свои первые шаги в изучении машинного зрения я начинал, как и все, окрылённый идеями Марра о «Начальном» или «Двух-с-половиною-мерном Эскизе», которые (согласно теории Марра) кратчайшим образом должны были привести нас к наилучшему описанию содержания картинки. (Под содержанием картинки подразумевалась, конечно, информация, содержащаяся в картинке).

«Видите, – азартно агитировал я своих начальников, стараясь убедить их выделить мне какие-то бюджетные средства и время на проведение нужных (по моим понятиям) исследований, – Вы видите, как несколькими штрихами, несколькими скупыми линиями художник передаёт вам полное представление о предмете, который он изобразил на картинке. Контурные и пограничные линии являются основными носителями информации в картинке. Научившись выделять и обрабатывать их, мы получим ключ к пониманию картинки!»

Мои начальники скептически ухмылялись и не спешили делиться со мной своими бюджетами. Однако, кое-что они всё-таки мне позволяли делать. И всё для того, чтобы я мог сам убедиться, что моя вера и мой оптимизм очень сильно преувеличены – вместо ключей к пониманию картинки я научился получать груду краевых элементов, выделяемых в результате прогона оператора 3x3 (или 5x5, или 7x7) по полю картинки в 256x256 пикселей. (В режиме почти реального времени.) Все силы и время уходили только на это. А в результате? Пшик, бесформенная груда краевых элементов (edges), которые невозможно ни сортировать, ни каким-либо образом группировать или связывать (помните «The binding problem»?), чтобы получить хоть какое-то подобие контура отдельного объекта или предмета.

Это был кошмар, и не только мой личный кошмар. Многие люди до сих пор барахтаются в этом болоте, не в силах освободиться от навязчивых идей о ценности пограничных линий. Многие до сих пор испытывают судьбу, пытаясь (надеясь) найти выход из положения. Во всяком случае, поток публикаций и победных реляций на эту тему (нашёл! нашёл!) не кончается по сей день (Ghosh et al., 2007; Awad & Man, 2008; Qiu & Sun, 2009).

Однако двадцать лет тому назад общая картина выглядела совсем не так уж мрачно. Хоть я вместе со всеми и суетился на «восходящем», «снизу-вверх» идущем магистральном пути по-пиксельной обработки картинок, мысли о заветной, «сверху-вниз» идущей, пиксель-связывающей идее не покидала меня никогда. Не очень точно представляя себе, что это значит, я всё таки учился и пытался извлекать из картинки Информацию.

А что это такое «информация»? Что такое «Визуальная» или «Зрительная» информация? – как и все мои сотоварищи (тогда и сейчас), я не очень точно представлял себе, что это такое. Мои учителя – Марр, например, который первым ввёл в употребление термин «зрительная информация» (Marr, 1978) – тоже не очень-то заботились о том, чтобы точно определить о чём же собственно говоря, идёт речь. Никому, правда, это никогда не мешало. И мне тоже. Я даже успел в эти годы изобрести для себя такое понятие, как «Информационное Содержание Отдельного Пикселя» (Single Pixel Information Content) и меру его количественного определения (Diamant, 2003). Экспериментируя с этой мерой, я изобрёл и научился вычислять «Удельное Информационное Содержание Картинки» (Image Specific Information Density) и неожиданно для себя обнаружил «Принцип Сохранения Информационного Содержания Картинки» (Image Information Content Conservation Principle). Согласно этому принципу, при уменьшении размера картинки, её удельное информационное содержание не меняется (до определённого предела), а в некоторых случаях даже чуть-чуть растёт. После достижения какого-то предела сжатия, её удельное информационное содержание резко падает. Вот этот пред-предельный размер картинки, считал я, и должен быть тем оптимальным размером, в котором должен осуществляться поиск информации в картинке. (Чтобы не тратить попусту силы и время на перебор всех пикселей в картинке. Помните, конечно, TigerSHARC?)

Своё замечательное открытие я, разумеется, попытался немедленно обнародовать, но, как обычно, был «решительно отвергнут». Одну из таких попыток – подачу на BMVC-2002 (British Machine Vision Conference, Британская Конференция по Машинному Зрению) – я бережно храню на своём сайте, и ссылка на него (Diamant, 2002) приведена в списке моих источников. Самое интересное во всей этой истории, что подобные работы были выполнены и подобные результаты были получены затем сотрудниками MIT – Массачусетского Технологического Института (Torralba, 2009). Правда, на семь лет позже. Правда, в результате психофизических экспериментов над людьми, а не количественных (как у меня) измерений.

Однако – «не к этому доводы». Главное, что мне удалось оторваться от господствующих традиционных представлений и продолжить развивать ту точку зрения, которая казалась мне более уместной. К этому времени было уже ясно, что поиск информации в картинке нужно начинать с её очень сжатого начального представления. К этому же подталкивали и известные экспериментальные данные, полученные в биологических исследованиях. Правда, если не слепо следовать за ними, а пытаться давать наблюдаемым фактам свои объяснения.

Дело в том, что технологи машинного зрения совсем не зря погнались за высокой разрешающей способностью своих фотоприёмников. От биологов им уже давно известно, что рассматривание наблюдаемой сцены человек осуществляет, последовательно сканируя её взглядом. При этом в каждый данный момент на нужный объект наводится центральная область глаза, так называемая фовея. (В отличие от фото- или теле-камеры, фотоприёмники (пиксели) человеческого глаза распределены по сетчатке глаза очень неравномерно. Очень высокая плотность фотоприёмников приходится на очень маленькую центральную область сетчатки, которая и называется фовеей.) Вот её-то, фовею, и направляет глаз (мозг) на рассматривание нужного предмета. Через неё и поступает к нам главная часть зрительной информации. По этой причине все исследования в области биологического зрения занимаются исключительно исследованием фовиального зрения, а все создатели электронных фотоприёмников стараются только его (фовиальное зрение) и имитировать.

Однако при всём этом один вопрос совершенно выпадает из обсуждения: если вся информация поступает к нам через фовею, каким образом глаз (мозг) знает, куда именно направить фовею в каждый данный момент? Ведь сканирование сцены осуществляется не по раз и навсегда заданному циклу, не по растру (как в телевидении, скажем), а как-то очень даже самопроизвольно, бессистемно. Откуда же мозг знает, куда именно должны смотреть глаза в каждый данный момент? Биологи не дают ответа на этот вопрос. (Впрочем, они его себе и не задают.)

Но возможный ответ на него мог бы быть: в мозгу у смотрящего наверняка есть общая карта «местности», по которой он решает, какой именно её участок ему следовало бы рассмотреть подробнее (куда направить свою фовею). Рассматривая и изучая сцену, человек одновременно пользуется двумя картами: крупного и мелкого масштаба. «Периферийные» области сетчатки обеспечивают его менее подробными картами, фовея – более подробными.

Между прочим, телевизионный оператор, камера которого имеет только фотоприёмник максимально возможного разрешения (чем-то напоминающий фовею), решает эту (возможно, классическую) общую для всех задачу с помощью трансфокатора – сначала снимается самый общий план, а потом камера «наезжает» на нужную часть сцены и сосредоточивается на её деталях.

Весь этот долгий реверанс в сторону биологического зрения нужен мне был только для того, чтобы ещё раз подчеркнуть уже напрашивающееся само собой: поиск информации в картинке должен начинаться с самой общей «карты местности», с самой сжатой и уменьшенной копии картинки.

Это, конечно же, противоречит всем известным теориям зрения, по которым поиск информации в картинке должен проводиться снизу-вверх, от пиксельных деталей к смысловому общему.

Поскольку одними только эмпирическими умозаключениями тут уже нельзя обойтись, нужно было срочно искать теорию, которая могла бы поддержать и укрепить это направление мысли.

Очень скоро оказалось, что такая теория есть. И даже не одна, а целых три – где-то в середине 1960-х годов прошлого столетия, приблизительно одновременно, но совершенно независимо друг от друга, были сделаны и опубликованы три работы, которые поначалу и не привлекли к себе особенного внимания: «Формальная теория индуктивного вывода» Р.Соломоноффа (Solomonoff, 1964), «Три подхода к численному определению информации» А.Колмогорова (Kolmogorov, 1965), и «О длине программ для расчёта конечных бинарных последовательностей» Г.Хаитина (Chaitin, 1966). Поскольку из всех трёх – Колмогоровская статья наиболее известна и популярна теперь, я буду в своих дальнейших рассуждениях ссылаться только на неё.

Как и Шенноновская Теория Информации (Shannon, 1948), опубликованная почти за двадцать лет до этого, Колмогоровская теория была направлена на поиск путей измерения количества «информации». Однако в то время как Шенноновская теория была занята оценкой среднего количества информации, полученной на выходе источника со случайным распределением сигнала, Колмогоровская теория была сосредоточена на информации, содержащейся в одном отдельном изолированном объекте. В моих глазах это гораздо больше подходило для обсуждения проблем, связанных с особенностями человеческого зрения.

Как и в случае с Марром и Анной Трейзман, я не стану утруждать моих читателей подробным изложением Колмогоровской Теории Сложности. Об этом вы можете и сами прочесть сегодня в замечательных текстах Пауля Витани, а также в трудах других многочисленных учеников и последователей А.Н.Колмогорова (Li & Vitanyi, 2008; Grunvald & Vitanyi, 2008). Мои намерения гораздо скромнее: опираясь на идеи Колмогоровской Сложности, попробовать взглянуть по-новому на проблемы человеческого зрения и поиска информации в картинках. Настойчиво повторяя при этом – Колмогоровская теория есть чисто математическая теория, ни о какой биологии не помышляющая и никакой биологией не озабоченная. В наши дни, когда любая идея в области машинного зрения или искусственного интеллекта спешит объявить себя «Biologically inspired» (до сих пор не могу найти достойного русского эквивалента этому термину – «биологией навеянный»? «биологией воодушевлённый»?), я повторяю опять и опять: «Искать объяснения нужно в чистой логике, в математике! Никакие «биологией навеянные» объяснения нам (вам) не помогут!..»

Как это всё отразилось на моих поисках информации в картинках и к чему всё это меня в конце-концов привело, вы узнаете, если у вас хватит терпения дочитать до конца эту статью.

## 5. Точка, точка, запятая – вышла рожица кривая...

Итак, первая и самая главная новость, которую я почерпнул для себя из теории Колмогорова, была: **информация – это описание**. Знаковое (буквенное, цифровое) или более сложное лингвистическое (языковое) описание, по которому предмет описания может быть достаточно точно восстановлен и воспроизведён. Колмогоров, а особенно его ученики, приравнивают такое описание к компьютерной программе, которая управляет реконструкцией исходного предмета описания (Vitanyi, 2006). В картинке такими предметами, безусловно, являются структуры, образованные конгломератами пикселей.

Колмогоровская теория указывает, как именно должны создаваться описания таких структур. Сначала создаются самые общие, упрощённые описания. На следующем этапе эти исходные начинают обрастать подробностями, становятся более детализированными. Этот процесс повторяется на всех последующих, низлежащих уровнях, таким образом, теоретически, весь процесс может идти бесконечно. Однако в случае зрения он идёт только до того уровня, где степень детализации обеспечивает системе принятие наиболее оптимального решения.

Этот подход очень напоминает другой давно известный принцип Оккамовой бритвы (Occam's Razor): «Из всех возможных предположений, описывающих конкретное наблюдение, всегда выбирай самое простое» (Sadrzadeh, 2008).

Таким образом, следуя Колмогоровской теории, мы можем заявить, что информационное описание, на самом деле, **это не какое-то единое монолитное описание, а целая иерархия описаний, где подробности и детали описания множатся на каждом более низком уровне иерархии**. Другими словами: начиная с самого обобщённого (упрощённого, сжатого) описания, иерархия информационного описания разворачивается «сверху-вниз», в соответствии с принципом «от-общего-к-деталям». (Внимание! Внимание, господа! Никаких «снизу-вверх» идущих потоков тут не упоминается, и ни о каких «проблемах связывания» тут нет и речи! Всё совсем наоборот! Вы понимаете, что это значит?!!)

А это значит, что реальный «сверху-вниз» идущий процесс извлечения информации из картинки обходится вообще без каких-либо предварительных знаний верхнего уровня! Потому, что знания верхнего уровня никакого отношения к извлечению информации из картинки не имеют! Потому, что её (информации верхнего уровня) там просто нет! Она есть и работает в голове у наблюдателя, который рассматривает картинку, но в самой картинке её нет. (Это было понятно ещё Спинозе, который почти триста лет тому назад сказал: «Красота – она в глазах смотрящего»).

Поняв это, я понял, что на самом деле **мы всегда имеем дело с двумя видами информации: объективной (физической) и субъективной (семантической).** Физическая информация содержится в картинке и есть описание структур, возникающих в массиве физических данных, представляющих данный объект. В нашем случае это пиксели, в других случаях это может быть всё, что угодно. Важно только, что это наблюдаемые структуры реальных (физических) данных, которые присутствуют в данном объекте. И поэтому **описание этих структур есть физическая информация, которая относится к (принадлежит) самому объекту**. Другой вид информации, с которым нам приходится иметь дело, это описание взаимоотношений и связей, имеющихся или могущих иметь место между отдельными элементами физических структур, наблюдаемых в этом объекте. Это всегда наше (человеческое) объяснение этих связей. Поэтому одна и та же картинка (одна и та же физическая информация) может быть по-разному понята (истолкована) разными людьми. То есть, одной и той же физической информации может быть присвоена разная семантика, разная семантическая информация, потому что эта информация принадлежит не самому объекту, а тому субъекту (зрителю, наблюдателю), который его (этот объект) рассматривает. То есть, **семантическая информация не содержится в картинке, а присваивается ей наблюдателем** (в его, наблюдателя, голове).

Это очень важное и замечательное наблюдение. Оно, конечно же, противоречит всем традиционным представлениям об информационном содержании картинки и способах его извлечения. Но оно не противоречит логике и здравому смыслу. Поэтому нужно было срочно проверить, как это выглядит на практике. Как можно, следуя предложенным мною принципам, провести сегментацию заданной картинки и получить конечный набор (нет, не готовых финальных объектов – это ведь работа для человеческого мозга), а «связанных» структурных субчастей этих (будущих) объектов.

Я проделал всю эту работу, и мне даже удалось опубликовать некоторые результаты моих экспериментов (Diamant, 2004; Diamant, 2005a; Diamant, 2005b). Я не буду пересказывать здесь их содержание (читатель всё это может прочесть и сам). Я хочу лишь указать на выводы, которые неизбежно следуют из всех этих экспериментов, пролагая путь к пониманию принципов работы человеческого зрения.

Во-первых, в картинке имеется только физическая информация, и только её можно извлечь из картинки. Процесс извлечения начинается со сжатия картинки до размера (приблизительно) в 100 пикселей. На этом уровне проводится начальная сегментация картинки, и начинается обратный процесс расширения картинки до её исходного (начального) размера. Процесс этот идёт ступенчато, на каждой более низкой ступени осуществляются коррекция средней интенсивности в каждом сегменте, последовательное уточнение границ сегмента и поиск зародышей новых (более мелких, внутренних) частей сегмента. На каждой ступени создаётся реестр описаний «объектов», выделенных на данной ступени. Каждое описание включает в себя: положение центра тяжести сегмента, площадь сегмента, длину пограничной линии, среднюю интенсивность (или цвет) сегмента, а также элементы топологии – «слева (справа) от сегмента АА», «выше (ниже) сегмента ВВ», «вложенная часть сегмента СС», и т.д. Всё вместе это представляет список «атрибутов» сегмента, т.е. список элементов описания сегмента, по которым, в конце концов, сам сегмент может быть восстановлен. Именно это я и называю физической информацией. Она всегда может быть извлечена из картинки, и никакой информации «верхнего уровня», никакой семантической информации для этого не требуется.

Семантическая информация, как любая другая информация (а теперь мы это уже твёрдо знаем), должна представлять собой иерархию описаний, где ближе к вершине располагаются более общие описания, а на низлежащих уровнях их всё более и более подробные детализировки.

Я уже сказал раньше (повторяя это вслед за Колмогоровым), что информационные описания должны быть всегда реализованы на каком-нибудь языке описания. Хотя сегодня существует уже очень много формальных языков описания, мне кажется, что для конструирования думающих машин, нам лучше всего было бы воспользоваться уже существующим языком человеческого общения. Ведь именно этим языком пользуются единственно знакомые нам и хорошо себя зарекомендовавшие разумные системы (системы человеческого разума).

В этом случае семантическая иерархия должна выглядеть следующим образом. Скажем, описание какого-нибудь акта в пьесе или сценарии обычно распадается на описания его отдельных сцен и мизансцен, а те, в свою очередь, на описания отдельных актёров и предметов, в каждой мизансцене участвующих. Деталировка описаний на этом не кончается, и каждый отдельный предмет в свою очередь может быть разложен на его составляющие компоненты.

Вот тут-то и вступает в игру физическая информация, потому что детали самого низкого уровня семантической иерархии это и есть та физическая информация, которую мы воспринимаем нашими органами чувств. (Эта физическая информация, как вы конечно помните, сама по себе есть разворачивающаяся сверху-вниз иерархия).

Обобщая сказанное, можно теперь уверенно заявить, что понимание физической информации, которую приносят нам наши органы чувств (зрения, например), есть встраивание (пристраивание) этой физической информации в соответствующую семантическую структуру, где такая (или очень близкая к ней) физическая информация уже присутствует. Мы ищем в своей памяти «рассказ» о предмете или событии, физическая информация которого наилучшим образом соответствует той физической информации, которую мы в данный момент наблюдаем. Этот извлечённый из памяти рассказ и есть интерпретация наблюдаемого в данную минуту. И если такого подходящего рассказа в нашей памяти не находится – мы не понимаем, что же мы видим, мы на самом деле просто ничего не видим. В биологии (психологии) зрения это известное явление, которое (перевожу, как умею, с английского) называется «Ситуационная Слепота» (Situation Blindness).

Этот поиск интерпретирующего рассказа в памяти человеческого наблюдателя может быть представлен алгоритмически, то есть может быть реализован машиной (компьютером). А это значит, что принцип работы думающих машин нам теперь ясен и мы можем немедленно приступить к их реализации. Осталось лишь несколько мелких деталей, которые нужно будет обсудить и решить уже по ходу дела... Однако, как говорится, «дьявол-то он всегда в деталях».

## 6. Что-то с памятью моей стало...

Изложенную только что принципиальную схему думающей машины я опубликовал уже несколько лет тому назад (Diamant, 2007; Diamant, 2008), но никакой реакции на это не последовало. Поэтому продолжу и дальше в гордом одиночестве...

Из всего сказанного выше однозначно следует, что основным элементом думающей машины (и разумного человека, конечно) является память, в которой накапливаются рассказы о всяких случаях из нашей прожитой жизни. Каждую новую физическую информацию, которую поставляют нам наши органы чувств, мы немедленно (и часто бессознательно) «пристраиваем» в один из имеющихся уже в нашей памяти рассказов, и если эта «пристройка» оказывается удачной, мы немедленно понимаем (узнаём) из текста рассказа, как же нам надлежит реагировать, то есть, что же нам делать дальше, ибо в тексте рассказа уже содержится описание всего того, «что было, что будет, и чем сердце успокоится». А тогда возникает законный вопрос: как же занести нам в память нашей машины все эти (на все случаи жизни подходящие) рассказы? С человеком, кажется, всё ясно – личный жизненный опыт и рассказы бывалых людей испокон веков составляли существенную часть его человеческого образования, т.е. заполнение его памяти нужными рассказами. Причём эти рассказы вовсе не должны быть связаны непосредственно с личным опытом конкретного индивида. А что же нам делать с машиной? Как её наделить человеческим опытом? Ответ на это очень прост – искусственно перенести (внести) в память машины те рассказы, которые будут нужны ей для решения того круга задач, которые мы хотели бы ей перепоручить.

Это, конечно, сильно снижает наши шансы создать человекоподобную разумную машину, но в этих своих мечтах мы очень часто и очень сильно перегибаем. Разве люди все одинаковы? Разве информация (рассказы) в памяти хирурга похожи на рассказы в памяти педиатра, или механика в авторемонтной мастерской, или шеф-повара в ресторане? Разве эти рассказы были получены ими из личного жизненного опыта? Конечно, нет! Они (люди) всему этому были научены, обучены. Эти знания (эта информация, эти рассказы) были ими получены уже в готовом виде, извне. Так почему же к машине мы должны относиться иначе? Почему от машины мы ждём универсальности и «самообучаемости», которые не существуют даже у людей?

Говоря это, я наживаю себе новых врагов, которые исповедуют и проповедуют сегодня теории о самообучении и самонаучении живых организмов в природе и интеллигентных машин в нашем распоряжении. Нет, – утверждаю я. – Никакого самообучения нет! Ни в природе, ни в искусственно создаваемых нами машинах. (То есть иногда это случается, но не как правило, а как редчайшее исключение. Как Ньютон или Эйнштейн, например, или какие-нибудь гениальные поэты. Все остальные получают свои знания всегда извне, всегда уже в готовом виде). Это хорошо видно в новейших исследованиях по обучению, т.е., передаче информации (знаний) от одного живого существа к другому (как среди бактерий так и среди высших животных) и целенаправленного обучения детёнышей в мире животных. (Я подробнее говорю об этом в своих других статьях, где цитирую соответствующие работы на эту тему). Это только у Киплинга Маугли ходит на двух ногах и разговаривает как человек, – в реальных же условиях дети, выросшие среди волков, например, могли передвигаться только на четвереньках и могли только повизгивать. Человеческому поведению их **не научили**. В то время как дети одной расы, выращенные в другой расовой среде, легко усваивают все правила поведения и язык этой среды. Поэтому нужно быстрее избавиться от веками освящённых предрассудков и быстрее пересаживать машинам ту часть человеческой памяти, те конкретные рассказы, необходимые им для решения весьма определённого круга задач, которые мы в каждом конкретном случае хотели бы им перепоручить (например, без устали смотреть в пустой экран систем видео-разведки и видео-наблюдения).

Очевидной новостью в этом процессе переноса знаний является то, что формой записи прошлого знания я считаю рассказ. (Надо было бы ещё до всего определить, что такое «знание». Не вдаваясь в подробное обсуждение этого весьма щекотливого вопроса, скажу только: для себя я определяю «знание» как «зафиксированную (записанную) информацию». Прошу именно это иметь в виду, когда вам встретится термин «знание»).

Очень интересный аргумент в пользу такого «литературного» («рассказ-ориентированного») подхода я нашёл недавно в работе Израиля Моисеевича Гельфанда и его соавторов, посвящённой опыту сотрудничества между врачами и математиками в деле создания «умной» машины для автоматического диагностирования заболеваний (Гельфанд и др., 1989). Создание такой машины потребовало особого языка общения между машиной и людьми. Обычно такие языки создаются в виде онтологического словаря терминов, одинаково понятных и машине, и людям, которые с ней работают. (Основополагающая статья на эту тему была опубликована Томасом Грубером только в 1993 году (Gruber, 1993). Гельфанд и его сотрудники столкнулись с этим и думали об этом ещё в 80-х.) (Кстати, Гельфанд – ученик А.Н.Колмогорова. Колмогоров принял его к себе в аспирантуру, хотя у Гельфанда не было даже соответствующего формального образования. Зато потом уже Академия Наук СССР отыгралась на нём за всё – его, почётного члена многих иностранных академий, советская Академия Наук отказывалась признать своим, и за границу представлять советскую науку его, конечно, не выпускали, пока не рухнула и сама империя, и её арийская наука).

Да, так вот что писал И.М.Гельфанд по поводу языка думающих машин: «...есть два способа развития (литературного) языка: написание высоко художественных произведений и составление толкового словаря. Мы знаем, какое большое влияние на развитие русского языка оказали и Пушкин, и Даль.» (Шекспир и Доктор Джонсон в английском языке, добавим мы от себя).

В те далёкие 80-е годы, создатели диагностирующей машины выбрали для себя путь создания словаря. Однако это не отменило и не отменяет справедливости первой части определения, данного И.М.Гельфандом. Смысл его прозрения стал мне понятен лишь после того, как я сам «дошёл» до рассказа как формы записи и представления семантической информации (Вот вам ещё один пример «ситуационной слепоты»). Остальной же мир по сей день так и занят составлением, сведением воедино и усовершенствованием онтологических словарей.

Говоря о «литературе» (о «рассказе») невозможно не задуматься о языке, на котором всё это должно быть записано. То, что информационное описание обязательно должно быть реализовано на каком-либо языке, совсем не мной придумано – об этом говорится ещё у Колмогорова. Нам же следует только внимательно проследить, куда нас могут завести все эти разговоры.

А ведут они нас вот куда: если верно то, о чём мы только что говорили, то память, в которой хранятся «рассказы» любого живого существа должна быть функционально очень похожа на память компьютера, т.е. быть такая, что в неё удобно вписываются (и из неё читаются) стринги соответствующих текстов. Есть такая память у живых существ? Конечно, да! И мы с ней хорошо знакомы – это геном, наша генетическая память. Хотя, если сравнивать с компьютером, то у последнего имеются два вида памяти: постоянная и оперативная (hardware and software). Если это правило не случайный артефакт, а действительно имеет всеобщий характер, тогда и у живых существ тоже должны иметься такие же два вида памяти: постоянная и оперативная. Постоянная ROM – Read Only Memory (только для чтения память) и оперативная RAM – Random Access Memory (Случайного обращения память). Так оно и есть! У каждого живого существа (в том числе и у человека), есть такая Постоянная память (только для чтения), которая уже давно всем известна и знакома – это наш геном, наша наследуемая генетическая память. У человека же есть (и каждый может подтвердить это на основании своего собственного опыта) и Оперативная память – то, что записывается и хранится у нас в голове в течение нашей индивидуальной жизни!

Я вижу, как грозно хмурятся мои оппоненты – ведь информация у нас в голове это совсем не стринги, а пачки электрических импульсов (так нас всегда учили нейробиологи), и хранится она совсем не в виде стрингов, а... Впрочем, как хранится у нас наша память никто не знает. Есть, говорят, у некоторых некие соображения, но всё это очень и очень туманно.

То, что любая информация, исходя из всего ранее сказанного в этом тексте, есть текстовое описание (а значит, стринги), оспаривать дальше уже бессмысленно. Что генетическая память – это набор химически записанных текстов (т.е. стрингов), уже давным-давно не новость. Всё это только ещё раз подтверждает вышесказанное. То, что в оперативной памяти мы, скорее всего, имеем дело с аналогичным явлением, т.е. информация между нейронами записывается и передаётся не как пачки импульсов, а как химически записанные фразы, – сегодня тоже уже известно многим и многим: астроциты, например, обмениваются информацией с нервными клетками без всяких импульсов (Voltera & Meldolesi, 2005; Haydon & Carmignoto, 2006). Именно поэтому их до недавнего времени в упор не видели – ведь искали, как всегда, Индию, и не хотели замечать лежащую перед ними Америку. «Ситуационная слепота», как водится.

«Но ведь электрические импульсы между нейронами – это реально наблюдаемый и измеряемый факт?!» – скажете вы. Верно, и наблюдаемый, и измеряемый. Но вот вам другой хорошо известный пример: во время работы струйного принтера из дюз его печатающей головки «выплёвываются» микродозы тонера (специальных чернил для принтера), которые собственно и образуют остающиеся на бумаге изображения. Вы можете ухитриться и замерить этот импульсный расход чернил, которые принтер тратит на печать заданного текста. А теперь решите: сможете ли вы когда-нибудь извлечь из этих измерений (импульсной мощности чернильных струй) смысл того, что вы только что отпечатали на бумаге? Вряд ли. Пачки импульсов, распространяющихся между нейронами, имеют такое же отношение к информации, как пачки импульсов тонера к напечатанному тексту.

К подобного рода примерам можно отнести и наблюдения за временными изменениями потока электронов в электронно-лучевой трубке осциллографа или телевизора. Хорошо было бы, если бы эти аргументы отложилось как-то в голове у моих читателей.

## 7. Хотел бы в единое слово...

Ввиду вышесказанного, мои отношения с профессиональными биологами (впрочем, как и с моими коллегами по компьютерному зрению) остаются весьма напряжёнными. А жаль. Мы могли бы быть друг другу полезны. Я имею в виду не только новую осмысленную интерпретацию той физической информации, которую биологи сегодня получают в громадных объёмах, но не могут её правильно истолковать, потому что путают физическую и семантическую информацию, не понимая (а потому не видя) разницы между ними. Есть ещё много других, более общих, более философских аспектов всего вышесказанного, которые нам хорошо было бы обсуждать совместно.

При этом я вовсе не собираюсь делать вид, что я не понимаю, как далеко мы отстоим друг от друга, что провозглашённый выше принцип «сверху-вниз» разворачивающегося информационного описания («от общего к деталям») прямо противоречит традиционному «снизу-вверх» идущему принципу собирания отдельных частных свидетельств, прежде чем кто-нибудь решится делать из них соответствующие выводы и обобщения. Я понимаю, что для многих это «сверху-вниз» идущее понимание и разумение может попахивать Божьим духом и провиденьем, но я уверяю вас, что это совсем не так.

На самом деле всё гораздо проще. Для практических целей совсем не нужно спрашивать себя всегда: «А как же всё это реализуется в природе?» (Богом, если хотите, или естественной эволюцией). Можно поставить вопрос и так, как это сделано в заглавии этой статьи: **А как должен быть устроен человеческий мозг**, чтобы его можно было бы воспроизвести и реализовать в машине? Такая постановка вопроса, безусловно, является еретической. И я не собираюсь делать вид, что я не понимаю этого.

К тому же, и как уже было заявлено выше, я вовсе не собираюсь делать вид, что я «biologically inspired» (биологией воодушевлённый), что я знаю или догадываюсь, как всё это устроено в природе. Совсем наоборот, я утверждаю, что я этого не знаю и знать не хочу. Я утверждаю, что природа не инженер, а значит никогда не придумывает и не изобретает ничего нового. Она приспосабливает и подгоняет то, что у неё уже есть под рукой. Иногда из этого получаются замечательные вещи. Но всегда на это уходит масса сил и времени, миллионы и миллиарды лет, которых в нашем распоряжении нет.

При этом самые главные вещи, которые определили и определяют судьбу *Homo sapiens* на этой планете, возникли не в результате естественной эволюции, а были именно изобретены, придуманы на пустом месте самим человеком, и естественных природных аналогов у них нет. Хотите примеры? Пожалуйста: прежде всего это целенаправленный труд и орудия труда. Затем использование огня для приготовления пищи (и вообще, использование вареной пищи). Затем – это наша разговорная речь, а потом уже и счёт, и письмо, в свой черёд. А дальше уже вообще пошло-поехало, масса всяких других изобретений, которые теперь называются культурой и без которых сегодня современного человека уже невозможно себе представить.

Тем не менее, я хотел бы вернуться и всё изложить по порядку, не выводя своих заключений из наблюдаемых в природе фактов, а отыскивая в массе известных уже экспериментальных данных подтверждение своим «сверху-вниз» идущим соображениям. При этом я исхожу из следующего предположения: если действительно высказываемые мною умозаключения не есть какие-то досужие измышления, а носят всеобщий и фундаментальный характер, то эти явления должны наблюдаться в природе и присутствовать на всех уровнях эволюционного развития живых существ от микроорганизмов и бактерий до высших приматов и человека.

Итак, первое утверждение: **информация есть лингвистическое описание**. Современные представления о генетическом коде очень хорошо согласуются с этим утверждением. Что нейроны в мозгу обмениваются между собой пачками электрических импульсов, которые есть информация, кажется, противоречит этому. Но это только кажется. Рассказ об астроцитах и безимпульсном обмене информацией между нейронами и астроцитами я уже приводил ранее, так же, как и соображения об отсутствии связи между нейронными импульсами и информацией. Повторяться не буду. Биологам придётся пересмотреть свои взгляды на этот счёт.

Без понимания информации как лингвистического описания невозможно понять разницу между физической и семантической информацией. **Физическая информация** это описания состояния внешнего мира, которые попадают к нам из внешнего мира через наши органы чувств. **Семантическая информация** это та интерпретация, которую получает у нас физическая информация, это тот рассказ из прошлой жизни (своей или чужой), в который поступившая физическая информация пристраивается, обретая таким образом смысл и основание для принятия последующих решений.

Для того, чтобы иметь возможность анализировать (пристраивать) вновь поступающую физическую информацию и соответствующим образом реагировать на неё, система (и любой живой организм, в том числе) должны обладать памятью, где должны храниться все эти нужные рассказы из прошлой жизни. При этом по аналогии с хорошо известными нам компьютерами я смею утверждать, что любая живая система должна обладать двумя видами памяти – жёсткой (или, как в компьютерах это называют, *hardwired*) и гибкой (или, как в компьютерах это называют, *softwired*) памятью. То есть, **наличие жёсткой и гибкой памяти** есть обязательное условие эффективной работы любой разумной системы. (В 6-ом разделе я называл эти два вида Генетической и Оперативной памятью. По смыслу это одно и то же, поэтому я и дальше буду пользоваться этими терминами попеременно).

Сказав это, нужно немедленно посмотреть, а что же мы имеем в этом смысле в биологии. С человеком тут всё понятно – наш генетический код, сохраняющийся веками, вполне подходит под определение жёсткой памяти. Память же, которая есть у нас в голове и которой мы денно и нощно пользуемся, хорошо укладывается в понятие гибкой памяти. А что же мы имеем в других, более низкого уровня биологических системах? У бактерий, например?

Я был удивлён, как много уже сделано в этом смысле (и исследовано, и опубликовано) в области обработки информации у бактерий. Израильтяне, оказывается, идут тут в первых рядах. Я привожу в списке источников несколько опубликованных совсем недавно замечательных работ Бен-Яакова и Яблонки. Некоторые из них ещё в процессе печати, и могут быть получены прямо с сайта Эшеля Бен-Яакова – http://star.tau.ac.il/~eshel/. (Конечно, и все другие цитируемые работы могут быть найдены там же).

Замечательны эти статьи тем, что несмотря на традиционные подходы и представления (которые неимоверно затрудняют правильное истолкование наблюдаемых явлений), их авторы сумели, тем не менее, правильно описать увиденное.

Дальше можно страницами цитировать из Бен-Яакова, но я отберу лишь самое существенное. Уже у бактерий действительно можно наблюдать раздельное существование жёсткой и гибкой памяти. Жесткая память – в хромосомах основного генома, гибкая память – в плазмидах. По Википедии, плазмиды – это фрагменты ДНК, расположенные в клетках вне хромосом. Могут встраиваться в основной геном, могут и «вырезаться» из основного генома и существовать (сосуществовать с ним), а также и вне его. Ben-Jacob считает их основным элементом, ответственным за связь между бактериями (как внутри одной колонии, так и между разными колониями) и за осуществляемый при этом перенос информации (Ben-Jacob, 2008). По Бен-Яакову именно так происходит Горизонтальный Перенос генов у бактерий, т.е. копирование из плазмида одной клетки в плазмид другой, а затем встраивание этого плазмида в основной геном, и уже потом передача новообретённой информации от родителей к детям традиционным путём, так называемым путём Вертикального переноса генетической информации. Именно так (по Бен-Яакову) и вырабатывается у бактерий резистивность к антибиотикам (Ben-Jacob et al., 2004).

Однако большая часть информации, перенесенной горизонтально между бактериями, не становится частью генетического кода, а продолжает оставаться в Оперативной памяти каждой бактерии (в её плазмидах). Это ведёт к образованию так называемой Коллективной памяти в колонии бактерий и явных элементов социального поведения у бактерий, что требует наличия общей семантики и постоянного общения между членами сообщества (Ben-Jacob, 2008). Всё это неопровержимо свидетельствует о том, что 1) у бактерий да существует Оперативная память, 2) хранится в ней семантическая информация, без которой немыслимы осмысленные коллективные действия, 3) эта семантическая информация есть продукт общения и обмена информацией между клетками, и 4) эта информация записана на определённом, всем клеткам понятном языке. Бен-Яаков отмечает даже существование отдельных диалектов этого языка и описывает, как по общему договору бактерии могут изменить существующий диалект, чтобы защитить себя от бактерий-отщепенцев, паразитирующих на трудах сообщества (Ben-Jacob, Shapira & Tauber, 2006).

Понятно, что существование оперативной памяти, обособленной и физически отделённой от генетической памяти, безусловно, должно было представлять эволюционное преимущество, так как обеспечивало большую гибкость и приспособляемость к условиям быстроизменяющейся внешней среды. Генетическая память должна аккумулировать более долгосрочный опыт. Она обязана быть более консервативной, чтобы обеспечить устойчивость вида. Оперативная же память нужна для кратковременной адаптации, хотя со временем она может переходить в устойчивую генетическую память (если изменившиеся условия жизни достаточно долго сохранялись во времени).

Понятно, что в процессе эволюции все элементы Оперативной памяти и её взаимодействия с Генетической памятью постоянно совершенствовались и усложнялись, достигнув своего предела и совершенства в сегодняшней человеческой памяти. Уже было показано, что даже на уровне бактерий семантическая информация поступает в оперативную память извне. А это в свою очередь предполагает наличие форм взаимосвязи и общения, а также обязательное наличие языка такого общения. Пусть очень примитивного, пусть самого что ни есть рудиментарного языка (описания и общения), но обязательно языка. Только на этом рудиментарном языке отдельные клетки и целые сообщества клеток могут обмениваться между собой информацией. Это должно быть особенно важно и требует постоянного подчёркивания, когда речь идёт о таком специальном сообществе клеток как мозг, которое уникально в том смысле, что предназначено только для обработки информации, поступающей от других клеточных сообществ (от органов чувств, например, живого организма). Естественно, обмен и накопление информации в оперативной памяти, ограниченные рамками обмена внутри одного живого организма, всегда будут очень убогими и скудными. Изобретение разговорного языка, а вслед за ним письменности и счёта, необычайно ускорили и интенсифицировали процесс обмена и накопления информации в оперативной памяти *Homo sapiens*. Но его базовые, фундаментальные принципы, безусловно, оставались неизменными. В этом смысле интересно было бы представить себе, как реализуются у нас в мозгу эти процессы записи и чтения в/из оперативной памяти. Поскольку биологи об этом ничего не знают, мы можем позволить себе опять порассуждать логически. Блок-схема и предполагаемый алгоритм процесса обработки семантической информации, которые мы описали ранее, ничего не говорят о том, как и откуда вносится в этот блок обработки иерархические представления семантической информации. Этим-то мы сейчас и займёмся.

Поскольку оперативная информация должна представлять собой текстовые стринги, расположенные вне основного генома, естественно предположить, что эти стринги хранятся в так называемых «дендритных шипиках» на дендритных ветвях нейрона и в синаптических капсулах на его аксонных отростках. Других возможных мест хранения этой информации я не вижу. (Совсем недавно, 17 Декабря 2009 года, Nature опубликовала статью (Yang et al, 2009), в которой описано размещение памятных записей в дендритных шипиках, наблюдавшееся *in vivo*, «в живую», на подопытных мышках. Какое замечательное и своевременное подтверждение моих спекуляций).

Поскольку на уровне бактерий информация, содержащаяся в операционной памяти (в плазмиде), и генетическая информация в хромосомах основного генома легко рекомбинируют, естественно предположить, что они записаны на одном и том же языке. Столь же естественно предположить, что в процессе эволюции язык этот видоизменялся. А поэтому в системе всегда существовали и существуют не один, а множество языков описания, так же, как и правила перевода (перехода) с одного языка на другой. Подтверждением этому могут служить хорошо известные процессы перекристализации записей краткосрочной памяти в записи долгосрочные, т.н. Долговременная Потенциализация (Long-Term Potentiation). Неизменными при этом должны были бы сохраняться лишь форма и структура записи, т.е. для каждой записи в оперативную память должна была бы быть синтезирована длинная последовательность молекулярных кодирующих элементов в виде двойной спирали ДНК. Мне не очень ясно пока, как именно синтезируется эта первоначальная запись, но мне кажется, что я хорошо понимаю, как должно происходить считывание информации из операционной памяти. Я утверждаю, что это должно происходить так же, как это происходит при репликации генов в генетической памяти. Т.е. двойная спираль расплетается, одна часть восстанавливается к своему первоначальному виду и сохраняется на месте (для последующего использования), а вторая часть (тоже восстановленая) «пускается в оборот», т.е. переносится в ту область мозга, где фактически осуществляется процесс поиска оптимального сочетания входной информации и соответствующего ей описывающего рассказа, хранящегося в операционной памяти.

Естественно предположить, что при восстановлении копируемой половинки записи, ещё до того, как она «пускается в оборот», происходит попутное «редактирование» этой записи (точно так же, как это происходит при репликации генетической информации), т.е. по каким-то неизвестным для нас правилам, происходит замена и перестановка отдельных букв, слов, целых фраз и даже отдельных кусков текста в первоначальной записи (Уотсон, 2008). Когда-то это называлось мутациями. Сегодня это называют редакциями.

Таким образом то, что когда-то было иллюзорным процессом мышления, сегодня материализуется в простой и вполне понятный алгоритм обработки информационных текстов.

## 8. Эпилог

Пора кончать. Рассказ о том, куда завело меня Колмогоровское понимание информации как лингвистического описания (или, короче говоря, текстовых стрингов), я, как умел, изложил и представил тут со всеми подробностями. Осталась одна маленькая деталь, один философский нюанс всей этой истории, который мне хотелось бы изложить хоть под занавес.

Оперативная информация (иерархия семантической и физической информации записанная в форме рассказа) занимает сегодня такое важное место в развитии вида *Homo sapiens*, что не грех было бы задуматься: А не кончилась ли для нас уже дарвиновская эволюция, уступив своё место Культурной эволюции, в которой мы теперь живём и которую вокруг себя наблюдаем?

Ограниченные возможности накопления информации в Оперативной памяти эволюционно более низких живых существ, сменились взрывным ростом возможностей человеческого разума, когда с изобретением разговорного языка, письменности и счёта стало обычным и доступным пополнение оперативной памяти отдельного индивидуума из резервуара внешней коллективной памяти (сначала ближайшего рода и племени, территориально обособленного народа, а затем, по мере изобретения новых средств хранения и переноса информации, расширение потенциального резервуара внешней коллективной памяти до размеров всего человечества).

Доступность информации из внешней коллективной памяти и возможность перезагрузки её в оперативную память отдельного индивида, а также эволюционные преимущества такого доступа сегодня столь очевидны, что нет смысла ещё подробнее останавливаться на этом. Нужно только отметить, что в свою очередь всё это ведёт к взрывному расширению объёма накопленной коллективной памяти человечества, к стремительной интенсификации возможностей коллективного человеческого разума.

Одним из аспектов этой интенсификации возможностей является то, что мы (человечество) получили возможность вмешиваться в сами результаты естественной эволюции – вмешиваться в генетическую память, свою и всех других живых существ. Вчера объявили о присуждении Нобелевской премии исследователям механизма деятельности теломеров. Генетически модифицированные продукты (Европа, правда, ещё отказывается их есть, но в Индии и Китае этот вопрос уже давно не стоит), исследование и лечение различных генетически обусловленных заболеваний, многочисленные примеры генной инженерии – всё это свидетельства того, как далеко мы уже ушли от времени дарвиновской эволюции и как влияет теперь на нашу жизнь культурная эволюция, которой мы обязаны сегодня буквально всем.

И еще одной вещи я хотел бы коснуться на прощание: Мемы. Придуманный Доукинзом эквивалент гена, ответственный за передачу культурной информации (Доукинз, 1976). Когда Чарльз Дарвин в 1859 году заканчивал писать своё «Происхождение видов» никакого понятия о механизмах наследственности ещё и в помине не было. Первые законы наследственности были открыты Менделем лишь в 1865 году. Тут же они были накрепко забыты, чтобы быть вновь переоткрытыми через тридцать пять лет, в 1900 году. Сам термин «ген» был введен в обращение только в 1909 году.

В 1953 году Уотсон и Крик открыли дву-спиральную структуру молекулы ДНК – физико-химической реализации живого гена. Таким образом почти сто лет теория эволюции существовала и развивалась независимо от новых идей и понятий, связанных с исследованием её основных материальных носителей. Поэтому, когда в середине 70-х годов прошлого столетия Ричард Доукинз сделал попытку дополнить Дарвина новейшими открытиями из области генетики, его книгу ждал ошеломительный успех. Она немедленно стала бестселлером, выдержала несколько изданий, самое последнее из которых появилось совсем недавно, в 2006 году.

Однако, несмотря на всю оригинальность и новизну подхода, Доукинз не мог разрешить одну из основных проблем эволюционного развития – как связаны между собой генотип и фенотип, как попадает к нам из нашего окружения культурная информация? Не углубляясь слишком далеко в подробности, скажу лишь, что для решения этой проблемы Доукинз изобрёл и ввёл в обращение понятие «мема» – некоего эквивалента и подобия гена, ответственного за передачу фенотипических и прочих культурных особенностей в процессе эволюции конкретного организма.

Идея была моментально подхвачена, и целая новая область знания, получившая название «Меметики», начала стремительно развиваться (Heylighen & Chielens, 2008).

Ни Доукинз, ни его многочисленные последователи не хотят замечать, как далеко завело их (и увело в сторону) их геноцентричное понимание эволюции, (т.е., что за всё, за всё происходящее с нами несут ответственность только наши гены). Ведь ген – это всего лишь долгосрочная (жёсткая) память с записанной в ней исторически неизменной информацией. А то, что Доукинз называет мемом, – это всего лишь оперативная (гибкая) память, в которую перегружается из коллективной памяти человечества то знание, которое данному конкретному инвивиду в данный конкретный момент обеспечивает наиболее оптимальные условия существования и выживания.

И всё это мы уже обсуждали выше. И нет смысла возвращаться этому опять и опять.

## 9. Источники

(Чтобы не перегружать статью слишком длинным списком источников, я привожу ниже ссылки только на те статьи, которые впервые встречаются лишь в этом тексте или представляют собой особый интерес. Все остальные ссылки могут быть найдены в моих ранее опубликованных статьях, которые доступны на сайте – http://www.vidia-mant.info).

Torralba, A. (2009). How many pixels make an image? *Visual Neuroscience*, Vol. 26, Issue 1, pp. 123-131, 2009. Available: http://web.mit.edu/torralba/www/.
Treisman, A. & Gelade, G. (1980). A feature-integration theory of attention, *Cognitive Psychology*, Vol. 12, pp. 97-136, Jan. 1980.
Ben-Jacob, E. (2008). Social behavior of bacteria: from physics to complex organization, *The European Physical Journal B*, vol. 65, pp.315-322, 2008.
Ben-Jacob, E.; Shapira, Y. & Tauber, A. (2006). Seeking the foundations of cognition in bacteria: From Schrodinger's negative entropy to latent information, *Physica A*, vol. 359, pp. 495-524, 2006.
Ben-Jacob, E. & Levine, H. (2006). Self-engineering capabilities of bacteria, *Journal of The Royal Society Interface,* vol. 3, no. 6, pp. 197-214, 22 February 2006.

Ben-Jacob, E.; Becker, I.; Shapira, Y. & Levine, H. (2004). Bacterial linguistic communication and social intelligence, *Trends in Microbiology*, vol. 12, no. 8, pp. 366-372, August 2004.

Jablonka, E. & Ginsburg, S. (2008). Epigenetic learning in non-neural organisms, *Journal of Bioscience*, vol. 33, no. 4, pp. XX-YY, October 2008.

Heylighen, F. & Chielens, K. (2008). Cultural Evolution and Memetics. In: B. Mayers (Ed.), Enciclopedia of Complexity and System Science, Springer-Verlag, Berlin Heidelberg, 2008. Available (Доступно на сайте): http://pespmc1.vub.ac.be/papers/PapersFH2.html.

Yang, G.; Pan, F. and Gan, W-B. (2009). Stably maintained dendritic spines are associated with lifelong memories, *Nature,* vol. 462, 17 December 2009, pp. 920-924.

Diamant, E. (2005). Searching for image information content, its discovery, extraction, and representation, *Journal of Electronic Imaging*, vol. 14, issue 1, January-March 2005.

Diamant, E. (2005a). Does a plane imitate a bird? Does computer vision have to follow biological paradigms?, In: De Gregorio, M., et al, (Eds.), *Brain, Vision, and Artificial Intelligence,* First International Symposium Proceedings. LNCS, vol. 3704, Springer-Verlag, pp. 108-115, 2005. Available: http://www.vidiamant.info.

Diamant, E. (2008). Unveiling the mystery of visual information processing in human brain, *Brain Research*, vol. 1225, pp. 171-178, August 15, 2008.

Ванюшин, Б.Ф. (2004). Материализация эпигенетики, или небольшие изменения с большими последствиями. Доступно: http://wsyachina.narod.ru/chemistry/apygenetics.html.

Репин, В.С. (2000). «Ковчег Жизни» на стапелях эволюции, *Новый Мир*, №12, 2000.

Джеймс Уотсон, (2008). ДНК и мозг: в поисках генов психических заболеваний, Публичные лекции фонда «Династия», Публичная лекция фонда «Династия», 3 июля 2008 года, Москва, Дом ученых, Доступно на сайте: http://elementy.ru/lib/lections.

Гельфанд, И.М., Розенфельд, Б.И., Шифрин, М.А., (1989). Очерки о совместной работе математиков и врачей, Издательство АН СССР Наука, Москва, 1989.

Ричард Доукинз, (1976). Эгоистичный ген. Доступно: http://fictionbook.ru/author/dokinz_richard/.